\documentclass[nohyperref]{article}

\usepackage{microtype}
\usepackage{graphicx}
\usepackage{subfigure}
\usepackage{booktabs} 

\usepackage{hyperref}

\usepackage[accepted]{icml2022}

\usepackage{amsmath}
\usepackage{amssymb}
\usepackage{mathtools}
\usepackage{amsthm}
\usepackage{multirow}
\usepackage{arydshln}
\usepackage[capitalize,noabbrev]{cleveref}

\theoremstyle{plain}

\theoremstyle{definition}

\theoremstyle{remark}

\usepackage[textsize=tiny]{todonotes}

\icmltitlerunning{Training Your Sparse Neural Network Better with Any Mask}

\begin{document}

\twocolumn[
\icmltitle{Training Your Sparse Neural Network Better with Any Mask}



\icmlsetsymbol{equal}{*}

\begin{icmlauthorlist}
\icmlauthor{Ajay Jaiswal}{yyy}
\icmlauthor{Haoyu Ma}{xxx}
\icmlauthor{Tianlong Chen}{yyy}
\icmlauthor{Ying Ding}{yyy}
\icmlauthor{Zhangyang Wang}{yyy}
\end{icmlauthorlist}

\icmlaffiliation{yyy}{The University of Texas at Austin}
\icmlaffiliation{xxx}{University of California, Irvine}

\icmlcorrespondingauthor{Zhangyang Wang}{atlaswang@utexas.edu}

\icmlkeywords{Machine Learning, ICML}

\vskip 0.3in
]


\printAffiliationsAndNotice{}  

\begin{abstract}
Pruning large neural networks to create high-quality, independently trainable sparse masks, which can maintain similar performance to their dense counterparts, is very desirable due to the reduced space and time complexity. As research effort is focused on increasingly sophisticated pruning methods that leads to sparse subnetworks trainable from the scratch, we argue for an orthogonal, under-explored theme: \textit{improving training techniques for pruned sub-networks, i.e. sparse training}. Apart from the popular belief that only the quality of sparse masks matters for sparse training, in this paper we demonstrate an alternative opportunity: one can \textit{carefully customize the sparse training techniques to deviate from the default dense network training protocols}, consisting of introducing ``ghost" neurons and skip connections at the early stage of training, and strategically modifying the initialization as well as labels. Our new sparse training recipe is generally applicable to improving training from scratch with various sparse masks. 
By adopting our newly curated techniques, we demonstrate significant performance gains across various popular datasets (CIFAR-10,  CIFAR-100,  TinyImageNet), architectures (ResNet-18/32/104, Vgg16, MobileNet), and sparse mask options (lottery ticket, SNIP/GRASP, SynFlow, or even randomly pruning), compared to the default training protocols, especially at high sparsity levels. 
{\small Code is at \url{https://github.com/VITA-Group/ToST}}.
\end{abstract}

\vspace{-0.5em}
\section{Introduction}
\vspace{-0.5em}
Deep neural networks (NN) have achieved significant progress in many tasks such as classification, detection, and segmentation. However, most existing models are computationally extensive and overparameterized, thus it is difficult to deploy these models in real-world devices. 
\begin{figure}
    \centering
    \includegraphics[width=9cm]{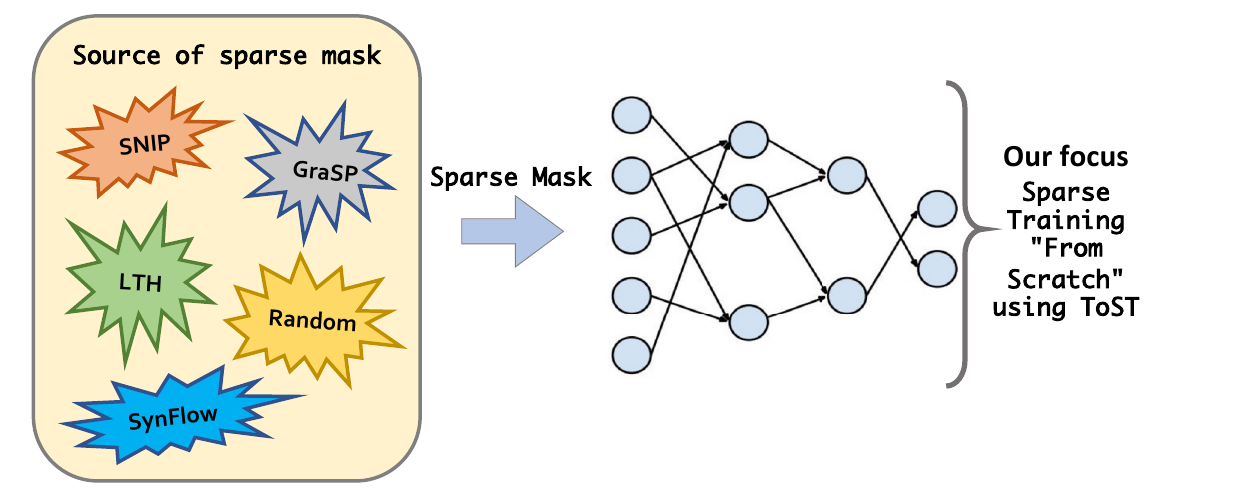}
    \vspace{-0.6cm}
    \caption{We aim to improve training of ANY sparse mask from scratch using our proposed sparse training toolkit (ToST).}
    \vspace{-0.6cm}
    \label{fig:basic_flow}
\end{figure} 
To address this issue, many efforts in direction of knowledge distillation \cite{Hinton2015DistillingTK}, quantization \cite{quantHilton2018}, and pruning \cite{LeCun1989OptimalBD}, have been devoted to compressing the heavy model into a lightweight counterpart. Among them, network pruning \citep{lecun1990optimal, han2015deep, han2015learning, li2016pruning, liu2018rethinking, frankle2018lottery}, which identifies sparse sub-networks (aka. sparse mask) by removing unnecessary connections, stands as one of the most effective methods. 

Recently, a significant amount of research efforts have been focused towards developing increasingly sophisticated and efficient pruning algorithms \cite{lee2018snip, wang2020picking, frankle2018lottery, frankle2019stabilizing, tanaka2020pruning}, to identify the sparse mask of the original dense model at the initialization, and then train only the sparse subnetwork from scratch, such as lottery ticket hypothesis (LTH) \cite{frankle2018lottery, frankle2019stabilizing}, SNIP \cite{lee2018snip}, GraSP \cite{wang2020picking}, SynFlow \cite{tanaka2020pruning}, and even random pruning \cite{su2020sanity,frankle2020pruning}. In most (if not all) cases, sparse masks obtained by various those pruning algorithms  are trained using the same training protocols optimized for dense neural network training. However, till now it is still \textit{under-explored and unclear} if dense training protocols are optimal for training a sparse mask from scratch too. We hence ask: \textit{Should training a sparse neural network requires atypical treatment and its own set of training toolkit or not?}

\begin{figure}
    \centering
    \includegraphics[width=8.5cm]{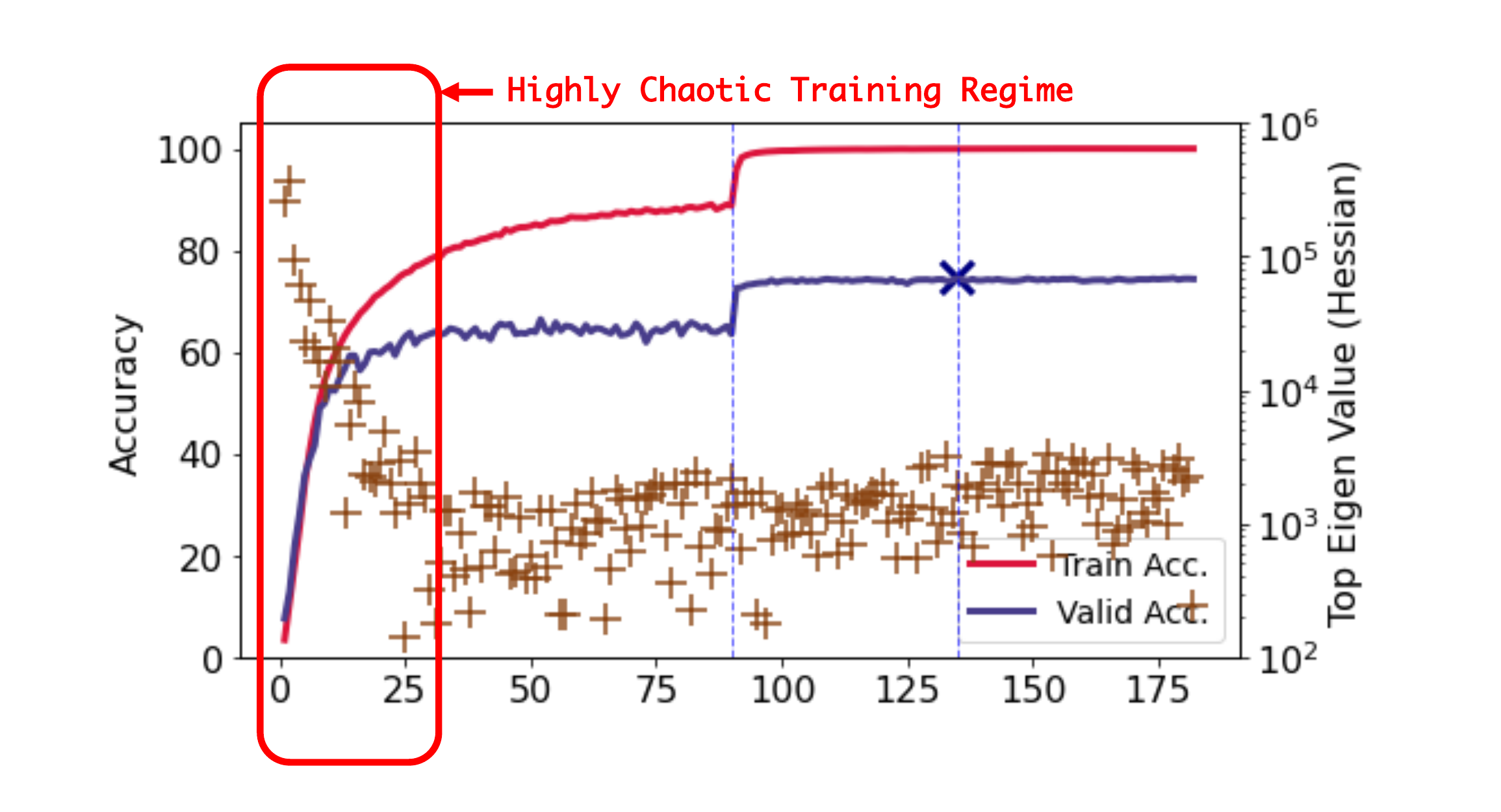}
    \vspace{-0.6cm}
    \caption{Top eigenvalues (Hessian) analysis of the training trajectory of a ResNet-18 sparse mask ($90\%$ sparsity) identified by LTH \cite{frankle2018lottery} using CIFAR-100.} 
    \vspace{-0.3cm}
    \label{fig:chaotic_reigm}
\end{figure}

Orthogonally to the popular belief that quality of sparse masks matter the most for sparse training from scratch (good sparse masks train better), this paper explores an under-explored and alternative opportunity: \textit{towards improving the training protocols of sparse sub-networks training.} This paper demonstrate that one can carefully customize the sparse training protocols to deviate from the default dense network training protocols, and achieve significantly better performance from the same sparse mask. Our work is primarily motivated by the observation thet sparse training suffers from poor gradient flow \cite{tessera2021keep}, and it has a highly chaotic optimization trajectory (Figure \ref{fig:chaotic_reigm}), which can lead to sub-optimal convergence of sparse masks. To tackle this problem, it is important to do a meticulous inspection of the training protocols of sparse training.

\paragraph{Our Contribution} We offer a toolkit of \textit{sparse retraining techniques} (ToST), and demonstrate that, solely by introducing our techniques into the training of sparse masks identified by existing popular pruning algorithms, our methods substantially reduce training instability, and improve performance and generalization of trained sub-networks. Our contributions can be summarized as:
\begin{itemize}
    \item In contrary to the common belief that quality of masks matter the most for sparse retraining, we argue for an orthogonal, and under-explored theme: \textit{improving the training techniques for pruned masks} and demonstrate that it significantly helps the sparse masks identified by various pruning algorithms to perform better. 
    \item We provide a curated and easily adaptable \textit{training toolkit (ToST)} for training ANY sparse mask from scratch: ``ghost" skip-connection (injecting additional non-existent skip-connections in the sparse masks), ``ghost" soft neurons (changing the ReLU neurons into smoother activation functions such as Swish \cite{ramachandran2017searching} and Mish \cite{misra2019mish}), as well as modifying initialization and labels.
    \item We report extensive experiments using variety of datasets, network architectures, and mask options. Incorporating our techniques in the sparse retraining immediately boosts the performance of sparse mask. This prompts to provide equal attention to improving the sparse retraining protocols rather than only focusing on designing better mask finding algorithms.
\end{itemize}

\vspace{-0.5em}
\section{Methodology}
\vspace{-0.5em}
In this section, we aim to provide a detailed introduction and motivation behind the tweaks in our sparse mask training toolkit (ToST). ToST consists of two \textbf{main tweaks}: ``Ghost" Soft Neurons (GSw), and ``Ghost" Skip Connections (Gsk), and some \textbf{miscellaneous tweaks} such as layer-wise re-scaled initialization, and label smoothing. We emphasize on highlighting the ``Ghostliness" behaviour of GSw and GSk, i.e, how they can be temporarily incorporated in sparse mask without making any final architecture changes.

\vspace{-0.5em}
\subsection{Revisiting Sparse Training}
\vspace{-0.5em}
Given a dense network $f(\boldsymbol{\theta}, \cdot)$, the sparse sub-network of it is defined as $f(\boldsymbol{\theta} \odot \boldsymbol{m}, \cdot)$, where $\boldsymbol{m} \in \{0,1\}^{\| \boldsymbol{\theta}\|_{0}}$ is the binary mask indicating the sparsity levels, and $\odot$ is the element-wise product.
$\boldsymbol{m}$ can be obtained from the pre-trained weights $\boldsymbol{\theta_d}$ or the random  initialization $\boldsymbol{\theta_0}$.  
The sparse training aims to train $f(\boldsymbol{\theta} \odot \boldsymbol{m}, \cdot)$ from scratch with training protocols $\mathcal{P}$. Previous works mainly focus on how to find a better $\boldsymbol{m}$ with $\theta_0$, while still apply the same $\mathcal{P}$ as the dense net $f(\boldsymbol{\theta}, \cdot)$.

\begin{figure}[h]
    \centering
    \includegraphics[width=7cm]{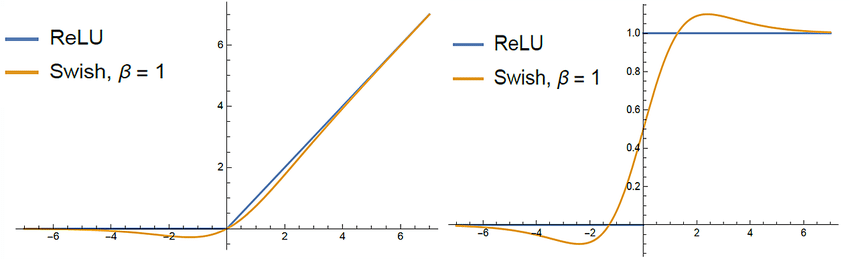}
    \vspace{-0.4em}
    \caption{ReLU and Swish (Parametric Swish with $\beta = 1$) activation functions along with their derivatives.} 
    \vspace{-0.8em}
    \label{fig:swish_and_relu}
\end{figure}

\vspace{-0.5em}
\subsection{``Ghost" Smooth Neurons and Skip Connection} 
\vspace{-0.3em}
\paragraph{Sparse Neural Networks and their trainability: } Highly sparse networks easily suffer from the layer-collapse \citep{tanaka2020pruning}, i.e., the premature pruning of an entire layer. This could make the sparse network untrainable, as the gradient cannot be backpropagated through that layer.  Additionally, in most (if not all) cases during the sparse neural network training, Rectified Linear Units (ReLU) \citep{nair2010rectified} are adapted as the default activation function ignoring the fact that ReLU is primarily optimized for training dense neural networks. However, the gradient of ReLU changes suddenly around zero (Figure \ref{fig:swish_and_relu}), and this non-smooth nature of ReLU is an obstacle to sparse retraining because it leads to high activation sparsity into the subnetwork (with many pruned weights), likely blocking a healthy gradient flow (Table \ref{tab:activation_sparsity}). During the training of sparse neural networks, we also observed that sparse training follows a highly chaotic optimization trajectory, which can lead to sub-optimal convergence of sparse subnetworks. 

\begin{table}[h!]
\centering
\begin{tabular}{lcccc} 
 \toprule
 \textbf{Activation} & \textbf{Layer 1} & \textbf{Layer 2} & \textbf{Layer 3} & \textbf{Layer 4}\\
 \midrule
 ReLU & 27.14\% & 39.33\% & 39.48\% & 57.93\%\\
 Swish & 0.31\% & 0.26\% &  0.24\% & 0.20\%\\ 
 Mish &  1.09\% & 1.14\% &  1.03\% & 0.95\%\\ 
 \bottomrule
     \vspace{-0.6em}
\end{tabular}

\caption{Layer-wise Activation sparsity of ResNet-18 sparse mask ($90\%$ sparsity) identified by LTH \cite{frankle2018lottery} and trained with CIFAR-100.}
\label{tab:activation_sparsity}
\end{table}
\paragraph{Injecting ``GSw" and ``GSk" in the sparse mask: } The non-smooth behavior of ReLU leads to high activation sparsity in the sparse network, and decreases its trainability by blocking the gradient flow. To mitigate this issue and encourage healthier gradient flow, we propose to temporally replace the ReLU to Swish \citep{ramachandran2017searching} and Mish \citep{misra2019mish} during the training of sparse masks. Different from ReLU, Swish and Mish are both smooth non-monotonic activation functions. The non-monotonic property allows for the gradient of small negative inputs, which leads to a more stable gradient flow \citep{tessera2021keep} during the training. 

\begin{figure}[h]
    \centering
    \vspace{-0.6em}
    \includegraphics[width=8.5cm]{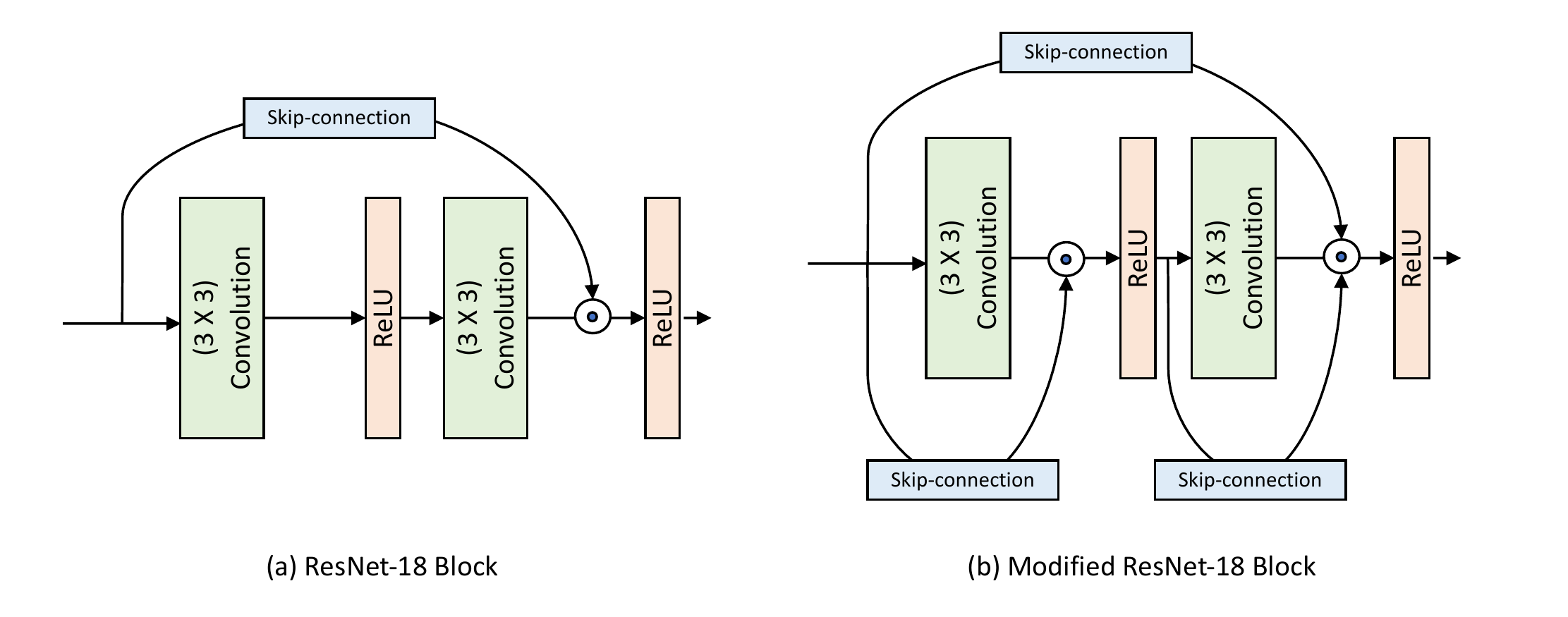}
\vspace{-1em}
    \caption{Our modified ResNet-18 block to introduce additional ``ghost" skip-connections for the initial stage of sparse training.} 
     \vspace{-0.6em}
    \label{fig:skip_model}
\end{figure}

With increase in sparsity, sparse neural networks suffers from layer-collapse \cite{tanaka2020pruning} which leads to blockage of gradient flow during training. The skip connection ( or named "residual-addition" ) \citep{He2016DeepRL} was initially proposed to avoid gradient vanishing problem, and enables the training of a very deep model. Motivated by the prevalent issue of dying kernels, high activation sparsity, and gradient blockage in highly sparse neural networks, we propose to ``artificially" inject temporal new skip-connections during the training. Figure \ref{fig:skip_model} illustrates this architectural modifications to the traditional Resnet-18 block. Similar to existing residual connection in traditional ResNet-18 block, our newly introduced skip-connections add input of each $(3 \times 3)$ convolution block, to their output before the activation. With high activation sparsity present in sparse subnetworks, additional skip-connections can facilitate healthy gradient flow and improve their trainability.

\label{section:ghostliness}
\paragraph{``Ghostliness" behaviour of ``GSw" and ``GSk": } Deep neural networks have been identified to learn ``low frequency features" initially \cite{rahaman2019spectral} roughly before the first learning rate annealing, followed by the next stage of learning high frequency features (later part of training). Both GSk and GSw are considered to primarily help the first stage, because Swish compared to ReLU alleviates aliasing at zero-truncation via a smoother decay window, while residual connections add more DC components and smoothen the loss landscape \cite{li2017visualizing}. During the low frequency learning stage (initial part of training), incorporating GSk and GSw will help maintain healthy gradient flow while focusing on the learning the low-frequency features. Considering their limited role in the second stage, it provide an opportunity to gracefully remove them while the training progresses. Note that during our experiments we observe that either their abrupt removal, or being removed too late in the training, will hurt the optimization adversely and lead to poor generalization.    

\begin{figure}
    \centering
    \includegraphics[width=7cm]{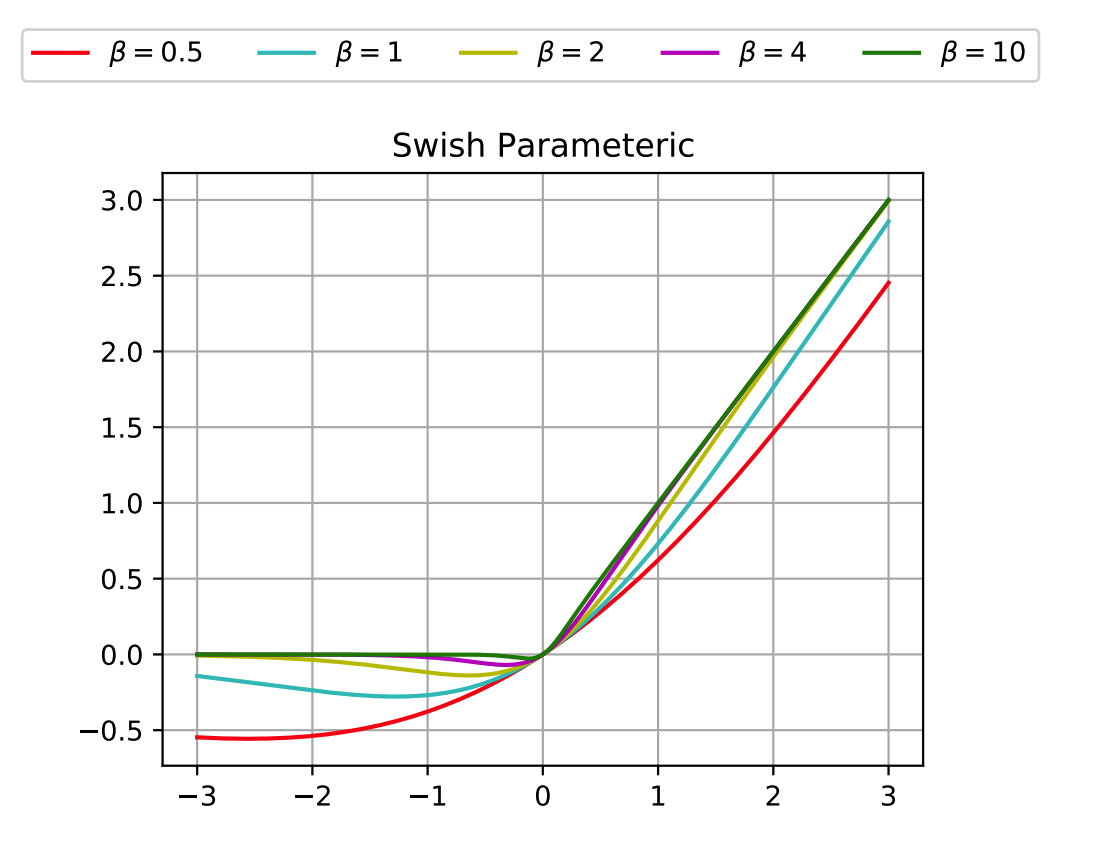}
    \vspace{-0.5cm}
    \caption{PSwish Visualization with different $\beta$ values.} 
     \vspace{-0.5cm}
    \label{fig:pswish}
\end{figure}


The parametric form of Swish (PSwish) function is $ f(x) = x \times sigmoid (\beta x)$. Note that  GSw is a special case of PSwish, when $\beta = 1$. PSwish transitions from identity function for $\beta = 0$, to ReLU for $\beta = \infty$ (smoothness decreases as $\beta$ increases). In our effort to keep the sparse mask architecture unchanged, we gradually increase the $\beta$ value of GSw, leading to be alike ReLU, and replace it with ReLU right before the first learning rate decay where the training regime changes \cite{leclerc2020two}. Following that, the training resumes as normal. 

Similarly, for GSk, we introduced gate functions regulated by a hyperparameter $\alpha$, which controls the contribution of GSk during the training. With $\alpha = 1$, GSk make full contribution during the training of the sparse mask. We decrease $\alpha$ in a scheduled way as training progresses, and finally set  $\alpha = 0$, which completely removes the role of GSk, right before the first learning rate decay. Note that ``Ghostliness" behaviour of GSw and GSk helps in reducing the additional training overhead, zeroing the inference overhead, and rehabilitating the original architecture of sparse mask.

One natural question which require attention following the ``Ghostliness" behaviour of GSw and GSk is: \textit{If we keep GSw and GSk throughout sparse training until the end, how that will impact their performance?} During our experiments, we found that keeping GSw and GSk forever during training provides slightly better performance. However, it will change the original backbone structure (hence unfair), and either nonlinear neurons or denser skip connections will add additional hardware latency during inference. In practice, it is viewed as a design trade-off for sparse neural networks; yet in this paper we stick to the same backbone architecture (including both neuron and skip pattern) as provided.

\vspace{-0.5em}
\subsection{Miscellaneous Tweaks}
\vspace{-0.5em}
\paragraph{Layer-wise Re-scaled initialization (LRsI): } Carefully crafted initializations that can prevent gradient explosion/vanishing in backpropagation have been important for the early success of feed-forward networks \citep{He2016DeepRL, Glorot2010UnderstandingTD}. Even with recent cleverly designed initialization rules, complex models with many layers and branches suffer from instability. For example, the Post-LN Transformer \citep{Vaswani2017AttentionIA} can not converge without learning rate warmup using the default initialization. 

In sparse subnetwork training, most existing works use common initializations \citep{Glorot2010UnderstandingTD, He2016DeepRL} directly inherited from dense NN (with the sparse mask applied). In sparse masks, the number of incoming/outgoing connections is not identical for all the neurons in the layer \citep{Evci2020GradientFI} and this raises direct concerns against the blind usage of dense network initialization for sparse subnetworks. Yet, \citep{Evci2020GradientFI} also showed that completely random re-initialization of sparse subnetworks can lead the sparse masks to converge to poorer solutions. 

To balance between these conflicting concerns, we propose to keep the original initialization of sparse masks intact for each parameter block and just re-scaled it by a learned scalar coefficient following recently proposed in  \citep{Zhu2021GradInitLT}. Aware of the sensitivity and negative impact of changing initialization identified by \citep{evci2020gradient}, we point that that linear scaling will \textbf{not} hurt the original sparse mask's initialization, thanks to the BatchNorm layer which will effectively absorb any linear scaling of the weights.  More specifically, we optimized a small set of scalar coefficients to make the first update step (e.g., using SGD) as effective as possible at lowering the training loss. After the scalar coefficients are learned, the original initialization of sparse mask is re-scaled and the optimization proceeds as normal. 

\vspace{-1em}
\paragraph{Label Smoothing (LS): }
Specifically, given the output probabilities $p_{k}$ from the network and the target $y_{k}$, a network trained with hard labels aims to minimize the cross-entropy loss by $L_{\text{LS}}= - \sum_{k=1}^{K} y_{k} \log \left(p_{k}\right)$, where $y_{k}$ is "1" for the correct class and "0" for others, and $K$ is the number of classes.  
Label smoothing \citep{szegedy2016rethinking}  changes the target to a mixture of hard labels with a uniform distribution, and minimizes the cross-entropy between the modified target $_{k}^{L S}$ and output $p_{k}$. The modified  target is defined as $y_{k}^{L S}=y_{k}(1-\alpha)+\alpha / K$, where $\alpha$ is the smooth ratio.  This uniform distribution introduces smoothness into the training and encourages small logit gaps. Thus, label smoothing results in better model calibration and prevents overconfident predictions \citep{muller2019does}. In our work, we propose to incorporate label smoothing during the training of sparse masks and show that it can effectively help in improving the performance of sparse masks. 
\vspace{-0.5em}
\section{Experiments and Analysis}
\subsection{Settings}
 Following the recent developments in pruning algorithms, we have used sparse masks identified from various pruning techniques: LTH \cite{frankle2018lottery}, SNIP \cite{lee2018snip}, GraSP \cite{wang2020picking}, SynFlow \cite{tanaka2020pruning}, as well as Random Pruning. Note that we have used the offical pytorch implementation of these algorithms to identify sparse mask, and train them with our ToST to evaluate the performance gain. For extensive validation across different datasets and architecture, we selected the most popular LTH \cite{frankle2018lottery}, and show how ToST generalizes across different datasets and architecture.

 In our experiments, all of our sparse masks has been trained using similar settings for simplicity in reproducing our results. For training, we adopt an SGD optimizer with momentum $0.9$ and weight decay $2$e$-4$. The initial learning rate is set to $0.1$, and the networks are trained for 180 epochs with a batch size of 128. The learning rate decays by a factor of $10$ at the $90$th and $135$th epoch during the training.  We run all our experiments 3 times to obtain more stable and reliable test accuracies. Note that apart from the tweaks proposed in ToST, we make no additional modification during the training process for fair evaluation. 

\begin{table*}
\centering
\small
\begin{tabular}{ccccccc} 
 \toprule
 \multirow{2}{*}{\textbf{Sparse Mask}}  & \multicolumn{3}{c}{\textbf{CIFAR-10}} & \multicolumn{3}{c}{\textbf{CIFAR-100}} \\ 
\cmidrule(rr){2-4}
\cmidrule(rr){5-7}
 & 90\% & 95\% & 98\%  & 90\% & 95\% & 98\% \\
 \midrule
 \textbf{ResNet-32 [No Pruning]}  & 94.80 & - & - & 74.64 & - & - \\
 \midrule
 Random Pruning & 89.95$\pm$0.23 & 89.68$\pm$0.15 & 86.13$\pm$0.25 & 63.13$\pm$2.94 & 64.55$\pm$0.32 & 19.83$\pm$3.21\\
 Random Pruning + ToST & \textbf{91.53$\pm$0.11} & \textbf{91.44$\pm$1.01} & \textbf{88.20$\pm$0.89} & \textbf{65.19$\pm$1.36} & \textbf{64.61$\pm$1.21} & \textbf{33.98$\pm$6.64}\\

 SNIP \cite{lee2018snip} & 92.26$\pm$0.32 & 91.18$\pm$0.17 & 87.78$\pm$0.16 & 69.31$\pm$0.52 & 65.63$\pm$0.15 & 55.70$\pm$1.13\\
 SNIP + ToST & \textbf{92.83$\pm$0.15} & \textbf{92.01$\pm$0.21} & \textbf{88.12$\pm$0.13} & \textbf{70.00$\pm$0.09} & \textbf{68.46$\pm$0.62} & \textbf{60.21$\pm$1.96}\\
 
 GraSP \cite{wang2020picking} & 92.20$\pm$0.31 & 91.39$\pm$0.25 & 88.70$\pm$0.42 & 69.24$\pm$0.24 & 66.50$\pm$0.11 & 58.43$\pm$0.43\\
 GraSP + ToST & \textbf{92.98$\pm$0.07} & \textbf{92.77$\pm$0.14} & \textbf{89.92$\pm$0.56} & \textbf{70.18$\pm$0.22} & \textbf{67.20$\pm$0.74} & \textbf{62.30$\pm$1.06}\\
 
 SynFlow \cite{tanaka2020pruning} & 92.01$\pm$0.22 & 91.67$\pm$0.17 & 88.10$\pm$0.25 & 69.03$\pm$0.20 & 65.23$\pm$0.31 & 58.73$\pm$0.30\\
 SynFlow + ToST & \textbf{93.39$\pm$0.59} & \textbf{92.06$\pm$0.32} & \textbf{91.82$\pm$0.73} & \textbf{70.25$\pm$0.06} & \textbf{67.90$\pm$1.22} & \textbf{61.72$\pm$0.84}\\
 
 LTH \cite{frankle2018lottery} & 93.14$\pm$0.30 & 92.98$\pm$0.12 & 92.22$\pm$0.61 & 71.11$\pm$0.57 & 70.37$\pm$0.19 & 69.02$\pm$0.22\\
 LTH + ToST & \textbf{94.01$\pm$0.23} & \textbf{93.60$\pm$0.70} & \textbf{93.34$\pm$1.06} & \textbf{72.30$\pm$0.61} & \textbf{71.99$\pm$0.95} & \textbf{70.22$\pm$0.61}\\
 
 \midrule
 
 \textbf{ResNet-50 [No Pruning]}  & 94.90 & - & - & 74.91 & - & - \\
 \midrule
 Random Pruning & 85.11$\pm$4.51 & 88.76$\pm$0.21 & 85.32$\pm$0.47 & 65.67$\pm$0.57 &  60.23$\pm$2.21  &  28.32$\pm$10.35\\
 Random Pruning + ToST & \textbf{92.73$\pm$0.22} & \textbf{90.95$\pm$1.22} & \textbf{87.11$\pm$2.21} & \textbf{67.75$\pm$1.32} & \textbf{63.60$\pm$0.11} & \textbf{41.99$\pm$4.51}\\

 SNIP \cite{lee2018snip} & 91.95$\pm$0.13  & 92.12$\pm$0.34  & 89.26$\pm$0.23  & 70.43$\pm$0.43  & 67.85$\pm$1.02  & 60.38$\pm$0.78\\
 SNIP + ToST & \textbf{92.89$\pm$0.53} & \textbf{92.56$\pm$0.12} & \textbf{90.56$\pm$0.19} & \textbf{70.79$\pm$0.22} & \textbf{68.06$\pm$0.09} & \textbf{61.51$\pm$1.41}\\
 
 GraSP \cite{wang2020picking} & 92.10$\pm$0.21  & 91.74$\pm$0.35  & 89.97$\pm$0.25  & 70.53$\pm$0.32 &  67.84$\pm$0.25 &  63.88$\pm$0.45\\
 GraSP + ToST & \textbf{92.64$\pm$0.17} & \textbf{92.33$\pm$0.09} & \textbf{90.94$\pm$0.35} & \textbf{70.89$\pm$0.21} & \textbf{68.09$\pm$0.12} & \textbf{65.01$\pm$0.33}\\
 
 SynFlow \cite{tanaka2020pruning} & 92.05$\pm$0.20 &  91.83$\pm$0.23  & 89.61$\pm$0.17 &  70.43$\pm$0.30 &  67.95$\pm$0.22 &  63.95$\pm$0.11\\
 SynFlow +ToST & \textbf{92.55$\pm$0.10} & \textbf{92.57$\pm$0.18} & \textbf{90.27$\pm$0.29} & \textbf{70.86$\pm$0.21} & \textbf{68.83$\pm$0.15} & \textbf{65.40$\pm$0.13}\\
 
 LTH \cite{frankle2018lottery} & 93.69$\pm$0.31 & 93.18$\pm$0.17 & 92.79$\pm$0.14 & 71.89$\pm$0.11 & 71.05$\pm$0.13 & 70.41$\pm$0.28\\
 LTH + ToST & \textbf{94.37$\pm$0.06} & \textbf{94.01$\pm$0.32} & \textbf{92.94$\pm$0.21} & \textbf{73.69$\pm$0.13} & \textbf{72.20$\pm$0.15} & \textbf{71.93$\pm$0.34}\\
 
\bottomrule
\end{tabular}
\caption{Classification accuracies of various pruning algorithm for varying sparsities $s \in \{90\%, 95\%, 98\%\}$ and network architectures (ResNet-18 and 32) with and without our sparse training toolkit (ToST).}
\vspace{-0.2cm}
\label{table:prune_methods}
\end{table*}

\subsection{ToST and ANY MASK} 
In this section, we conduct a systematic study to understand the performance gain by our proposed sparse training toolkit (ToST), when they are incorprated in the training process of ANY sparse identified by various pruning algorithms. Table \ref{table:prune_methods} demonstrate the effectiveness of our proposed toolkit on the sparse masks obtained by recently proposed pruning algorithms: SNIP \cite{lee2018snip}, which is a sensitivity based pruning algorithm, GraSP \cite{wang2020picking}, which is a Hessian based pruning algorithm, SynFlow \cite{tanaka2020pruning}, which is an iterative data-agnostic pruning algorithm, Lottery Ticket \cite{frankle2018lottery}, which is based on iterative magnitude pruning, as well as Random Pruning . We have used CIFAR-10 and CIFAR-100 during the evaluation of our ToST on various sparse masks obtained by pruning a substantial amount of parameters of ResNet-18 and ResNet-50 with varying sparsities $s \in \{90\%, 95\%, 98\%\}$. 

The results are summarized in Table  \ref{table:prune_methods}. We first observe that among all the pruning methods, sparse masks obtainned by LTH \cite{frankle2018lottery} perform the best at high sparsity for both CIFAR-10 and CIFAR-100. In comparison, sparse mask trained with ToST stays stable in performing significantly better across all pruning methods, datasets, and network architectures. Very interestingly, we observe that ToST can help randomly pruned masks at $98\%$ sparsity to achieve up to $\sim14\%$ and $\sim13\%$ (CIFAR-100) better results, for ResNet-32 and ResNet-50 respectively. This provided a strong indication towards the training stability provided by ToST during the sparse training even the mask quality is not great. Similarly for GraSP mask with 98\% sparsity, ToST provides $\sim4\%$ improvement. 

We additionally evaluated SNIP and LTH sparse masks with sparsities $s \in \{85\%, 90\%, 95\%\}$ on \textbf{TinyImageNet} \cite{deng2009imagenet}. Table \ref{table:tinyimagenet} presents the summary of our results. Similar to our results on CIFAR-10 and CIFAR-100, ToST provided sufficient performance boost to ResNet-50 sparse masks identified by SNIP and LTH, on the  larger TinyImageNet dataset. At $95\%$ sparsity, it provides $>2\%$ improvement for SNIP, and $>1.5\%$ improvement for the LTH mask. These benefits prompt a greater potential to reconsider the exploration and provide attention to improving sparse retraining strategies.

\begin{table*}
\centering
\begin{tabular}{lccc} 
 \toprule
 \textbf{Algorithm} & \textbf{85\%} & \textbf{90\%} & \textbf{95\%} \\
 \midrule
 SNIP \cite{lee2018snip} & 58.91$\pm$0.23 & 56.15$\pm$0.31 & 51.19$\pm$0.47\\
 SNIP + ToST  & 59.44$\pm$0.09  & 57.19$\pm$0.21 & 53.21$\pm$0.08 \\
 LTH \cite{frankle2018lottery} & 60.11$\pm$0.13 & 58.46$\pm$0.17 & 53.19$\pm$0.31\\
 LTH + ToST & 61.52$\pm$0.32 & 58.96$\pm$0.08 & 54.76$\pm$0.22\\
 \bottomrule
\end{tabular}
\caption{Classification accuracies on TinyImageNet for varying sparsities  $s \in \{90\%, 95\%, 98\%\}$ using ResNet-50.}
\label{table:tinyimagenet}
\end{table*}

\begin{table*}
\centering
\begin{tabular}{lccccc} 
 \toprule
 \textbf{Method} & \textbf{75\%} & \textbf{80\%} & \textbf{85\%} & \textbf{90\%} & \textbf{95\%} \\
 \midrule
 LTH \cite{frankle2018lottery} &  73.21$\pm$0.17 & 72.94$\pm$0.12 & 71.91$\pm$0.22 & 71.12$\pm$0.30 & 69.57$\pm$0.19\\
 LTH + GSk  & 73.77$\pm$0.11 & \textbf{73.69$\pm$0.25} & 72.86$\pm$0.30 & \textbf{72.17$\pm$0.23} & \textbf{71.72$\pm$0.22}\\
 LTH + GSw & 73.45$\pm$0.13 & 73.22$\pm$0.43 & \textbf{73.27$\pm$0.31} & 72.03$\pm$0.12 & 70.85$\pm$0.52\\
 LTH + LRsI & \textbf{73.93$\pm$0.15} & 73.12$\pm$0.13 & 72.30$\pm$0.19 & 71.83$\pm$0.32 & 69.98$\pm$0.29\\
 LTH + LS & 73.58$\pm$0.28 & 73.70$\pm$0.32 & 72.65$\pm$0.25 & 71.93$\pm$0.20 & 70.19$\pm$0.14\\
 \midrule
 LTH + ToST & 74.29$\pm$0.31 & 74.03$\pm$0.14 & 73.90$\pm$0.49 & 73.23$\pm$0.27 & 72.08$\pm$0.10\\
 \bottomrule
\end{tabular}
\caption{Breakdown of the performance of individual tweaks in ToST tweaks when applied on training ResNet-18 sparse masks (LTH) with varying sparsities $s \in \{75\%, 80\%, 85\%, 90\%, 95\%\}$ and trained on CIFAR-100.}
\label{table:performance_breakdown}
\end{table*}

\vspace{-0.5em}
\subsection{Performance Breakdown of ToST}
\vspace{-0.5em}
Our toolkit (ToST) consists of two \textbf{main tweaks}: ``Ghost" Soft Neurons (GSw), and ``Ghost" Skip Connections (Gsk), and some \textbf{miscellaneous tweaks} such as layer-wise re-scaled initialization, and label smoothing. While these tweaks when jointly applied in the training of sparse masks, significantly provides huge performance gain (Table \ref{table:prune_methods}, \ref{table:tinyimagenet}), an obvious question is: \textit{How our proposed tweaks helps in performance of sparse masks, when they are applied in isolation?} To answer this question, we selected LTH masks (considering it high popularity and better performance at high sparsity) with varying sparsities $s \in \{75\%, 80\%, 85\%, 90\%, 95\%\}$  for detailed evaluation of our individual tweaks.

Table \ref{table:performance_breakdown} summarizes the performance comparison of our individual tweaks when they are applied in isolation during the training of sparse masks. We can observe that ``GSk" standalone is the most effective tweak in improving the performance at very high sparsity with a performance gain of $2.15\%$ at $95\%$ sparsity. ``GSw" stands out to be the second most effective tweak in our toolkit for high sparsity, with performance gain of $1.36\%$ and $1.28\%$ at sparsity level $85\%$ and $95\%$ respectively. It is worth noticing that ``LRsI" achieve highest performance gain at $75\%$ sparsity which hints that each tweak helps in the trainability of sparse networks in its own unique way. When we combine these tweaks together to train the sparse LTH tickets, we observe the overall performance boost is significantly better than the individual tweaks. We get $1.08\% - 2.51\%$ performance gain within the sparsity range of $s \in [75-95]\%$. This clearly highlights the orthogonal benefits of our tweaks in sparse mask training.

\vspace{-0.5em}
\subsection{``Ghostliness" of GSw and GSk}
\vspace{-0.5em}
As discussed in Section \ref{section:ghostliness}, ``GSk" and ``GSw" primarily help in first stage of learning, we are motivated to remove them gradually during the course of time (aka. ``ghostliness"). Gradual removal will help in reducing the additional training overhead, zeroing the inference overhead, and rehabilitating the original architecture of sparse mask. To investigate the impact of ``ghostliness" behaviour, and how it may impact in unleashing the true strength of ``GSk" and ``GSw", when they are kept throughout sparse training until the end, we attempted to compare the performance with and without ``ghostliness" of our tweaks. 

\begin{figure*}
    \centering
    \includegraphics[width=14cm]{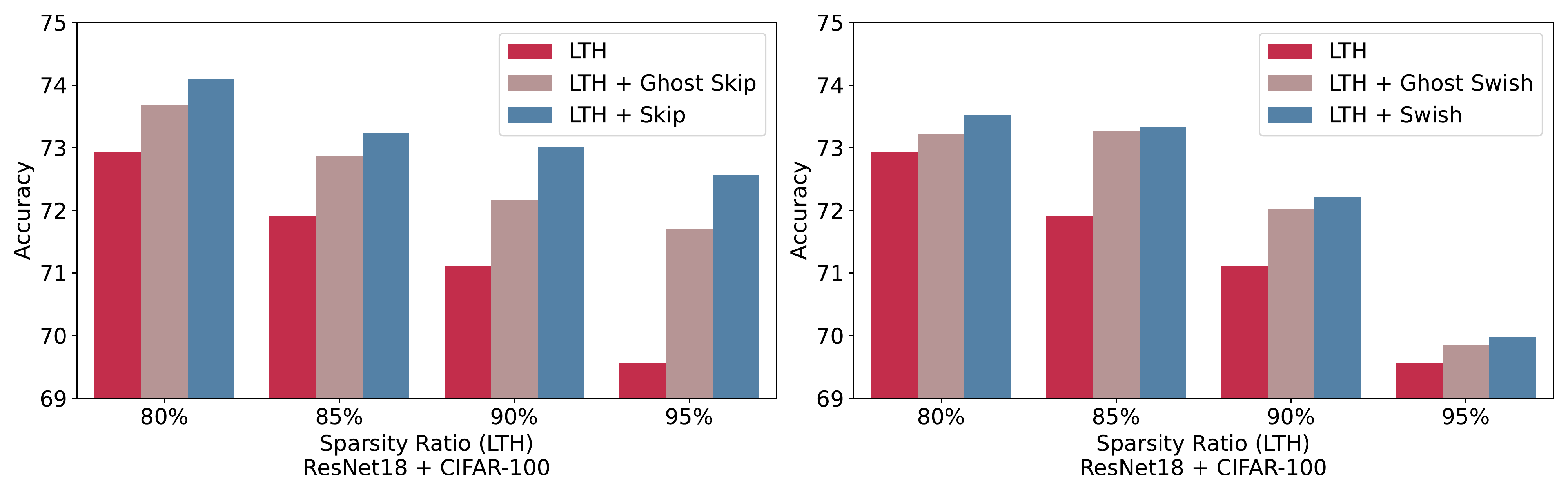}
    \caption{Performance comparison of the ``Ghostiliness" behaviour of GSk and GSw with the default prolonged injection of swish and skip connections for LTH sparse masks with varying sparsities $s \in \{80\%, 85\%, 90\%, 95\%\}$. } 
    \label{fig:ghost}
\end{figure*}

Figure \ref{fig:ghost} summarizes the performance comparison of the ``Ghostiliness" behaviour of GSk and GSw with the default prolonged injection of swish and skip connections for LTH sparse masks with varying sparsities $s \in \{80\%, 85\%, 90\%, 95\%\}$. We observed that keeping the skip connections, and swish throughout sparse training until the end provides some additional performance benefit (marginal for swish), but it comes up at the cost of additional hardware latency during the inference time. In practice, we identify this as a design trade-off for the sparse neural networks. To complete the analysis, we attempted to analyse mask performance if we ghost GSk and GSw after the first learning rate decay, and we found that the performance decreases by $-0.917\%$ and $-0.429\%$ (95\% sparsity) for GSk and GSw respectively, compare to our proposed settings. Additionally, abrupt removal of GSk and GSw towards the end of training, leads to significant performance drop of $>1.2\%$ (sparsity 95\%) for both GSk and GSw.   


\begin{figure*}
    \centering
    \includegraphics[width=\linewidth]{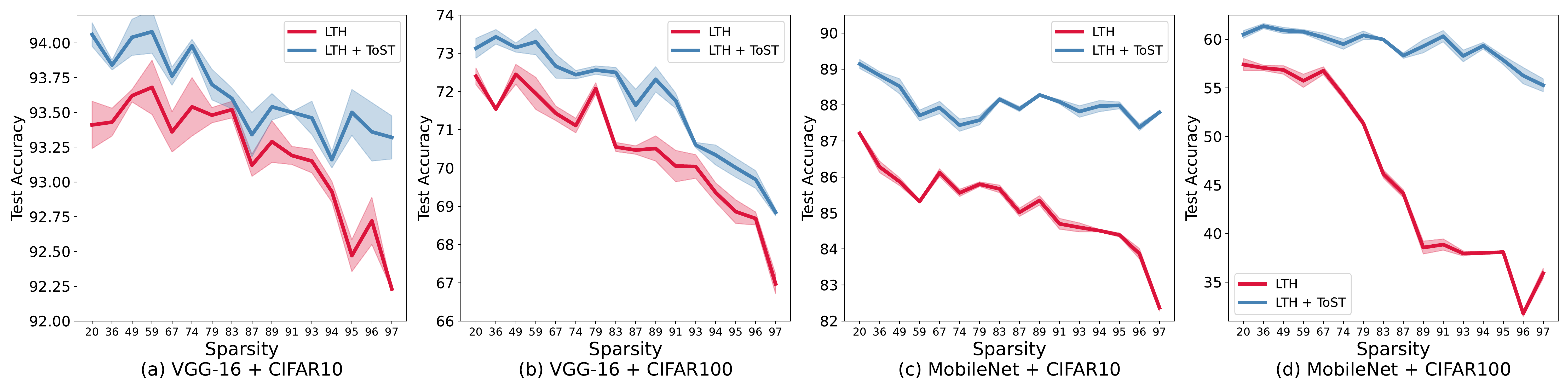}
    \vspace{-0.5cm}
    \caption{Performance comparison of sparse masks by LTH at varying sparsities $s \in [20\%-97\%]$ on CIFAR-10 and CIFAR-100.} 
    \label{fig:model_ablation}
\end{figure*}

\vspace{-0.5em}
\section{Ablation and Analysis}
\subsection{Generalization across Datasets and Architectures} 
\vspace{-0.5em}
In this section, we additionally evaluate the performance of ToST on VGG-16 \cite{simonyan2014very} and MobileNet \cite{howard2017mobilenets} using CIFAR-10 and CIFAR-100. Note that we have mainly studied LTH \cite{frankle2018lottery} masks hereinafter for ablations, considering their superior performance in comparison to SNIP, SynFlow, GraSP, and Random Pruning. Figure \ref{fig:model_ablation} summarizes the performance comparison of our sparse training toolkit when it is used to train the LTH sparse tickets with sparsity ranging from $s \in [20\%-97\%]$. In the plot, the red line indicates the performance of sparse tickets when they are trained using default setting proposed in \cite{frankle2018lottery, frankle2019stabilizing} while the blue line indicates the sparse tickets when trained using our ToST without any other additional modification for fair comparison. Clearly, our proposed tweaks help significantly in improving the performance of sparse tickets across all sparsities. Moreover, it is important to observe that the performance benefits of our tweaks increases significantly with increase in the sparsity level. This observation augment the necessity of ToST, while training sparse subnetworks with high sparsity.


\vspace{-0.5em}
\subsection{Smoothness of Loss Landscape} 
\vspace{-0.5em}
In this section, we try to understand the implications of our techniques during training of sparse subnetworks, through some common lens. Our methods can be viewed as a form of \textit{learned smoothening} \cite{chen2020robust} which is incorporated at an early training stage. Smoothening tools can be applied on the logits (naive label-smoothening \cite{label-smoothening}, knowledge distillation \cite{hinton2015distilling}), on the weight dynamics (stochastic weight averaging \cite{izmailov2018averaging}), or on regularizing the end solution.

\begin{figure}[h]
    \centering
    \vspace{-0.6em}
    \includegraphics[width=5cm, trim= 40 0 0 0]{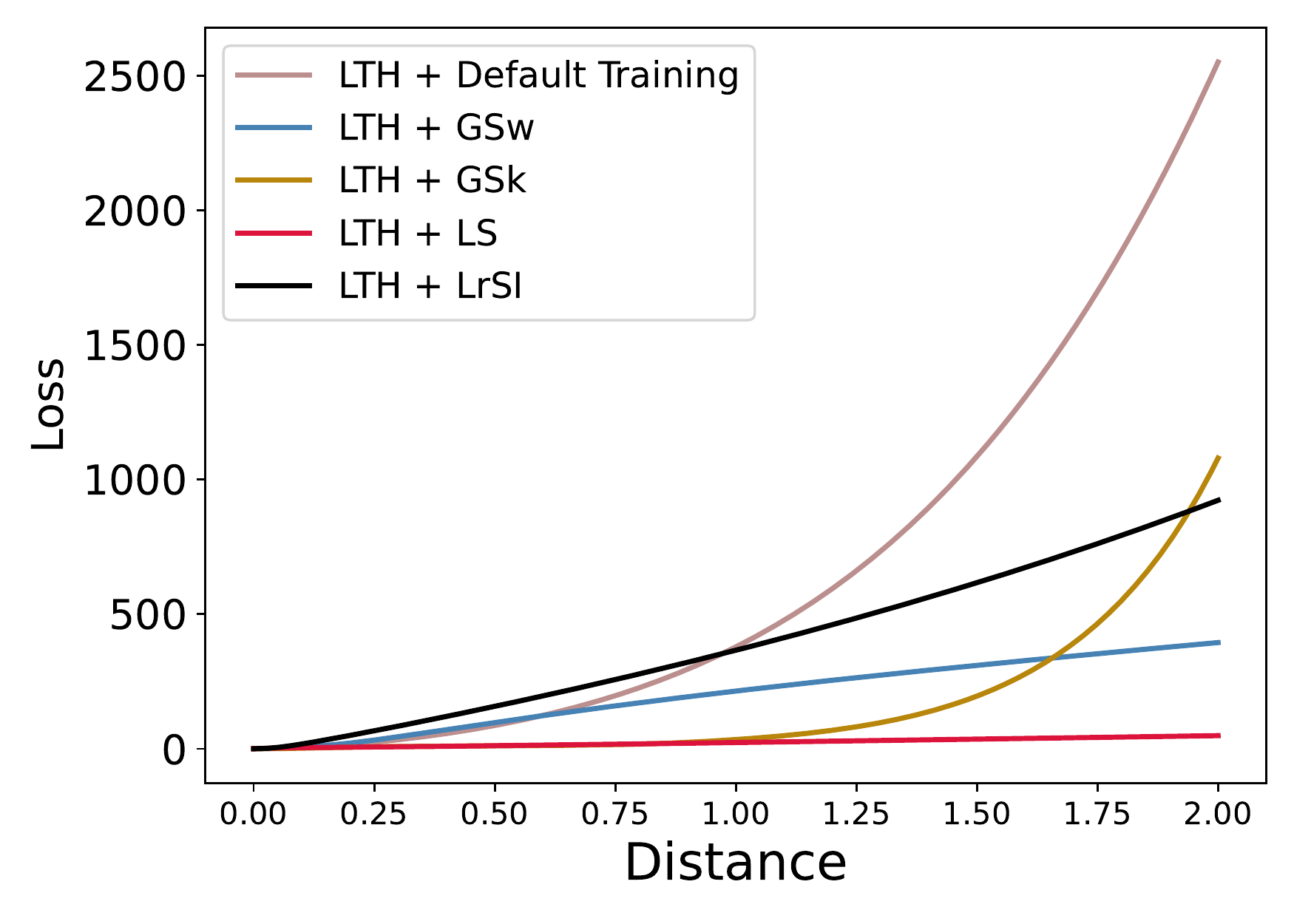}
    \caption{The change in testing loss as a function of perturbed weight distance, in  the  direction  of  top  eigenvector  of Hessian  matrix \cite{Yao2020PyHessianNN} of LTH ticket (90\% sparsity) for  ResNet-18  trained on CIFAR-100.} 

    \label{fig:smooth2}
\end{figure}

\begin{figure*}
\centering
\includegraphics[width=\linewidth]{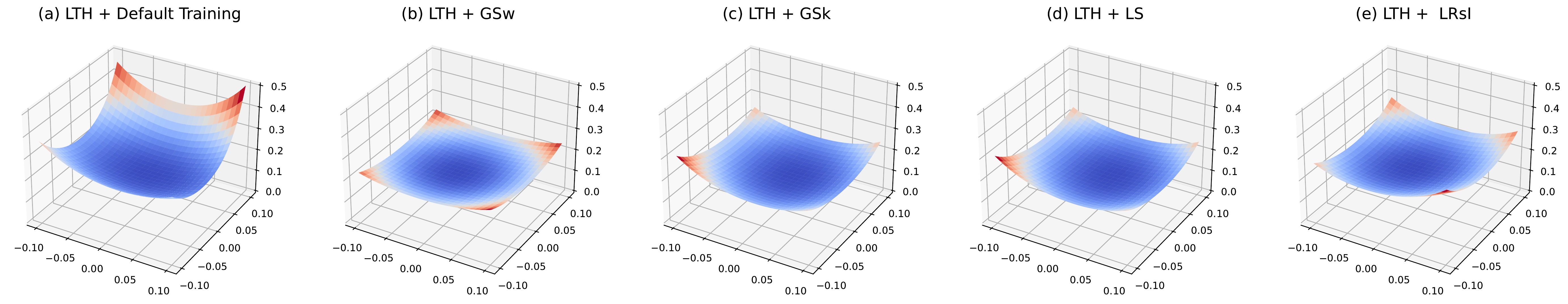}
\vspace{-1em}
\caption{Comparison of loss landscape of LTH ticket (90\% sparsity) from Resnet-18 trained using default dense training protocols proposed in \cite{frankle2018lottery, frankle2019stabilizing} and individual tweaks in our sparse toolkit ToST. Loss plots are generated with the same original images randomly chosen from CIFAR-100 test dataset using \citep{loss-landscape}. z-axis denote the loss value. }
\label{fig:loss_landscape}
\end{figure*}

We expect tweaks in ToST to find flatter minima for sparse mask training to improve its generalization, and we show it to indeed happen by visualizing the loss landscape w.r.t both input and weight spaces. Figure \ref{fig:loss_landscape} shows the comparison of loss landscape of LTH ticket (90\% sparsity) from Resnet-18 trained using default dense training protocols proposed in \cite{frankle2018lottery, frankle2019stabilizing} and individual tweaks in our sparse toolkit ToST.  It can be observed that each one of our tweaks notably flatten the sharp landscape w.r.t. the input space, compare to the default baseline of using \cite{frankle2018lottery, frankle2019stabilizing}, which aligns with our hypothesis that our tweaks can be viewed as some form of ``learned smoothening".  

Figure \ref{fig:smooth2} follows \cite{Yao2020PyHessianNN} to perturb the trained LTH sparse mask (90\% sparsity) in weight space, to show the flattening effect of our tweaks. It shows the change in testing loss as a function of perturbed weight space, in the direction of top eigenvector of Hessian obtained by \cite{Yao2020PyHessianNN}. Our methods present better weight smoothness around the achieved local minima, which suggests improved generalization \cite{Dinh2017SharpMC, Petzka2019ARF}.

\begin{figure*}
\centering
\includegraphics[width=12cm]{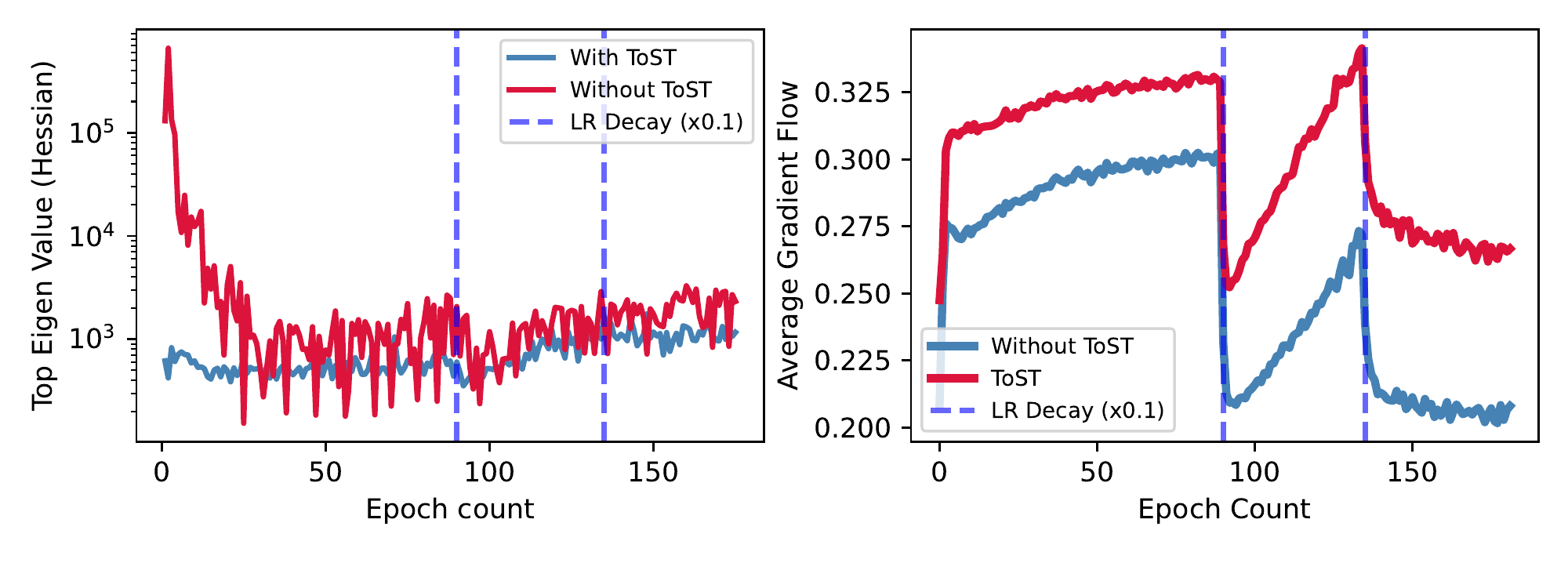}
\vspace{-1.7em}
\caption{(a) Comparison of Top eigenvalues (Hessian) of training trajectory of a ResNet-18 sparse mask (90\% sparsity by LTH) on CIFAR-100 with and without ToST. (b) Comparison of Average Gradient Flow \cite{tessera2021keep} ResNet-18 sparse mask (90\% sparsity by LTH) on CIFAR-100 during training with and without ToST. }
\label{fig:rebuttal}
\end{figure*}

\begin{table}[h]
\centering
\small
\begin{tabular}{lcccc} 
 \toprule
  & \textbf{Dense NN (0\%)} & \textbf{20\%} & \textbf{75\%} & \textbf{95\%} \\
 \midrule
 \textbf{``GSk"} & -0.77\% & +0.03\% & +0.56\%& +2.15\%\\
 \textbf{``GSw"} & +0.11\%& +0.29\% & +0.24\%& +1.28\%\\
 \bottomrule
\end{tabular}
\caption{Performance benefit of ``GSk" and ``GSW" when applied to dense networks (0\%) sparsity, low sparsity (20\%), mid-level sparsity (75\%), and high sparsity (95\%). We have used LTH sparse mask of ResNet-18 trained on CIFAR-100.}
\label{table:dense_vs_sparse}
\end{table}
\subsection{Are ``GSk" and ``GSw" same helpful in Dense NNs?}\vspace{-0.3em}
In this section we attempt to answer one important question: \textit{How does ``GSw" and ``GSk" impact the performance of dense network? Are they equally beneficial to training dense networks too?} Table \ref{table:dense_vs_sparse} illustrates the performance benefits of GSw and GSk when they are applied at various level of sparsity ranging from 0\% (corresponds to dense network) to low-level (20\%), mid-level (75\%), and finally high-level (95\%). It clearly answer aforementioned question that both ``GSk" and ``GSw" significantly help the sparse networks more than the dense network and the performance benefits enlarges with increasing sparsity. 

\textbf{Remark:} The recently proposed RepVGG \cite{ding2021repvgg} cleverly adds skip-connections (SKs) to dense networks of \textit{VGG-like plain topology} (no SKs) by re-parameterization during training, and later removing SKs at inference. In contrast, our ``GSk" is applied to training sparse networks of \textit{arbitrary topology},  mostly ResNets with pre-existing native SKs. Our experiments further reveal that adding extra SKs can even \textbf{hurt} the performance of dense ResNets with pre-existing SKs (e.g., ``-0.77\%" in Table \ref{table:dense_vs_sparse} for ResNet-18). Meanwhile, adding SKs during sparse training of those ResNets, using our proposed soft alternative, benefits their performance consistently, especially at high sparsity. \textit{Our lesson is}: sparsity has an overlooked important role in influencing whether more skip connections will benefit, potentially due to the trade-off between network representation capacity and optimization easiness.

\subsection{Effect of ToST on Hessian and Gradient Flow} \vspace{-0.3em} Hessian eigenvalue/spectral density \cite{Yao2020PyHessianNN} can be used to analyze the the topology of the loss landscape, and its magnitude indicates the ``degree of smoothness in loss landscape" and the ease for Stochastic Gradient Descent to converge to a good solution. Higher value of top eigenvalue indicate poorer quality of loss landscape and difficult optimization. Figure \ref{fig:rebuttal}(a) illustrates the effect of ToST on top eigenvalues of Hessian for the training trajectory of ResNet-18 sparse LTH mask with 90\% sparsity. Furthermore, to effectively measure the gradient changes before and after ToST, we calculated the \textit{Average Gradient Flow} for the unpruned weights during training \cite{tessera2021keep}. Figure \ref{fig:rebuttal}(b) presents the comparison of the gradient flow with and without ToST when training ResNet-18 sparse LTH mask with 90\% sparsity. Clearly, ToST facilitates healthier gradient flow during the training of sparse neural networks. 

\section{Related Work}
\paragraph{Network Pruning }
Pruning is fruitful in reducing network inference costs. 
In general, there are two types of pruning: One is unstructured pruning, which usually removes redundant weights. The important score of weights can be obtained from magnitude \cite{han2015deep, han2015learning}, gradient \cite{molchanov2017variational,molchanov2019importance} or Hessian \cite{lecun1990optimal}. The other is structured pruning, which prunes the entire channels or layers \cite{liu2017learning, li2016pruning, wen2016learning, he2017channel}. All of them starts with the fully trained dense model, and finetune the sparse network to achieve similar accuracy. 

\paragraph{Sparse Training}
Sparse training aims to train a sparse network from scratch. It can be categorized into two groups: \textcircled{1} \textit{Static sparse training}, which prunes the network at the initialization and maintains the pruning mask throughout training. Lottery Ticket Hypothesis (LTH) \cite{frankle2018lottery, frankle2019stabilizing,evci2019difficulty,savarese2020winning,chen2020lottery2,gale2019state,chen2020lottery} suggests that a dense network contains several sparse sub-network that can match the accuracy of the original model when trained in isolation from scratch. Later on, the Single-Shot Network Pruning (SNIP) \cite{lee2018snip} uses the gradients of the training loss at initialization to prune the network. The Gradient Signal Preservation (GraSP) \cite{wang2020picking} prune connections based on the gradient flow. The Iterative Synaptic Flow Pruning (SynFlow) \cite{tanaka2020pruning} preserves the total flow of synaptic strengths through the network to handle the layer-collapse issue. \cite{sung2021training} selects the sparse mask via Fisher information. \textcircled{2} \textit{Dynamic sparse training}, which allows the pruning mask to be updated during the training. They usually prunes weights based on the magnitude and grows weights back \cite{mocanu2018scalable} at random or based on the gradient \cite{evci2020rigging,liu2021we,chen2022sparsity,chen2021chasing}.  
All of these works imply that the quality of pruning mask is vital in sparse training. 

\cite{lee2019signal} analyzed sparse subnetworks from the signal propagation perspective and proposes a new technique of re-fitting initialization to improve their trainablity. More specifically, provided with sparse topology $C$ and initial random weights $W$, \cite{lee2019signal} optimizes  $W \rightarrow W^*$ such that the combination of the sparse topology and weights become layerwise orthogonal. In comparison, our LRsI technique keeps the original sparse weight initialization, except learning a small set of scaling coefficients per block to improve the gradient propagation. Along with being computationally more efficient, it can also better preserve the \textit{good initialization} already found in some sparse mask schemes such as LTH, which have been confirmed as necessary for their success \cite{tessera2021keep}.

\paragraph{Smoothness in Neural Network} 
Infusing smoothness into neural networks, including on the weights, logits, or training trajectory, is a common techniques to improve the generalization and optimization \cite{jean1994weight}. For labels, smoothness is usually introduced by replacing the hard target with soft labels \cite{szegedy2016rethinking} or soft logits \cite{hinton2015distilling}. This uncertainty of labels helps to alleviate the overconfidence and improves the generalization. 
Smoothness can also implemented by replacing the activation functions \cite{misra2019mish, ramachandran2017searching}, adding skip-connections in NNs \cite{He2016DeepRL}, or averaging along the trajectory of gradient descent \cite{izmailov2018averaging}. These methods contribute to more stable gradient flows \cite{tessera2021keep}  and smoother loss landscapes, but most of them have not been considered nor validated on sparse NNs.

\section{Conclusion}
This paper takes one step towards improving the training techniques for sparse neural networks. Contrary to the popular belief that only the  quality  of  sparse  masks  matters  for  sparse training, this paper presents an alternative opportunity that one can carefully customize the sparse training techniques to train sparse sub-networks identified by various pruning algorithms, and achieve significant performance benefits. It presents a curated and easily adaptable training toolkit for training any sparse mask from scratch, without any additional overhead. Extensive experiments across different pruning algorithms, sparse masks, and datasets shows the effectiveness of the proposed toolkit. Our future work will aim
for more theoretical understanding of the role of our toolkit in sparse training performance improvement.

\section*{Acknowledgement}
Z.W. is in part supported by an NSF RTML project (\#2053279).

\bibliography{example_paper}

\begin{thebibliography}{60}
\providecommand{\natexlab}[1]{#1}
\providecommand{\url}[1]{\texttt{#1}}
\expandafter\ifx\csname urlstyle\endcsname\relax
  \providecommand{\doi}[1]{doi: #1}\else
  \providecommand{\doi}{doi: \begingroup \urlstyle{rm}\Url}\fi

\bibitem[Chen et~al.(2020{\natexlab{a}})Chen, Frankle, Chang, Liu, Zhang,
  Carbin, and Wang]{chen2020lottery2}
Chen, T., Frankle, J., Chang, S., Liu, S., Zhang, Y., Carbin, M., and Wang, Z.
\newblock The lottery tickets hypothesis for supervised and self-supervised
  pre-training in computer vision models.
\newblock \emph{arXiv preprint arXiv:2012.06908}, 2020{\natexlab{a}}.

\bibitem[Chen et~al.(2020{\natexlab{b}})Chen, Frankle, Chang, Liu, Zhang, Wang,
  and Carbin]{chen2020lottery}
Chen, T., Frankle, J., Chang, S., Liu, S., Zhang, Y., Wang, Z., and Carbin, M.
\newblock The lottery ticket hypothesis for pre-trained bert networks.
\newblock \emph{arXiv}, abs/2007.12223, 2020{\natexlab{b}}.

\bibitem[Chen et~al.(2021{\natexlab{a}})Chen, Cheng, Gan, Yuan, Zhang, and
  Wang]{chen2021chasing}
Chen, T., Cheng, Y., Gan, Z., Yuan, L., Zhang, L., and Wang, Z.
\newblock Chasing sparsity in vision transformers: An end-to-end exploration.
\newblock \emph{Advances in Neural Information Processing Systems},
  34:\penalty0 19974--19988, 2021{\natexlab{a}}.

\bibitem[Chen et~al.(2021{\natexlab{b}})Chen, Zhang, Liu, Chang, and
  Wang]{chen2020robust}
Chen, T., Zhang, Z., Liu, S., Chang, S., and Wang, Z.
\newblock Robust overfitting may be mitigated by properly learned smoothening.
\newblock In \emph{International Conference on Learning Representations},
  2021{\natexlab{b}}.

\bibitem[Chen et~al.(2022)Chen, Zhang, pengjun wang, Balachandra, Ma, Wang, and
  Wang]{chen2022sparsity}
Chen, T., Zhang, Z., pengjun wang, Balachandra, S., Ma, H., Wang, Z., and Wang,
  Z.
\newblock Sparsity winning twice: Better robust generalization from more
  efficient training.
\newblock In \emph{International Conference on Learning Representations}, 2022.
\newblock URL \url{https://openreview.net/forum?id=SYuJXrXq8tw}.

\bibitem[Deng et~al.(2009)Deng, Dong, Socher, Li, Li, and
  Fei-Fei]{deng2009imagenet}
Deng, J., Dong, W., Socher, R., Li, L.-J., Li, K., and Fei-Fei, L.
\newblock Imagenet: A large-scale hierarchical image database.
\newblock In \emph{2009 IEEE conference on computer vision and pattern
  recognition}, pp.\  248--255. Ieee, 2009.

\bibitem[Ding et~al.(2021)Ding, Zhang, Ma, Han, Ding, and Sun]{ding2021repvgg}
Ding, X., Zhang, X., Ma, N., Han, J., Ding, G., and Sun, J.
\newblock Repvgg: Making vgg-style convnets great again.
\newblock In \emph{Proceedings of the IEEE/CVF Conference on Computer Vision
  and Pattern Recognition}, pp.\  13733--13742, 2021.

\bibitem[Dinh et~al.(2017)Dinh, Pascanu, Bengio, and Bengio]{Dinh2017SharpMC}
Dinh, L., Pascanu, R., Bengio, S., and Bengio, Y.
\newblock Sharp minima can generalize for deep nets.
\newblock In \emph{ICML}, 2017.

\bibitem[Evci et~al.(2019)Evci, Pedregosa, Gomez, and
  Elsen]{evci2019difficulty}
Evci, U., Pedregosa, F., Gomez, A., and Elsen, E.
\newblock The difficulty of training sparse neural networks.
\newblock \emph{arXiv preprint arXiv:1906.10732}, 2019.

\bibitem[Evci et~al.(2020{\natexlab{a}})Evci, Gale, Menick, Castro, and
  Elsen]{evci2020rigging}
Evci, U., Gale, T., Menick, J., Castro, P.~S., and Elsen, E.
\newblock Rigging the lottery: Making all tickets winners.
\newblock In \emph{International Conference on Machine Learning}, pp.\
  2943--2952. PMLR, 2020{\natexlab{a}}.

\bibitem[Evci et~al.(2020{\natexlab{b}})Evci, Ioannou, Keskin, and
  Dauphin]{Evci2020GradientFI}
Evci, U., Ioannou, Y.~A., Keskin, C., and Dauphin, Y.
\newblock Gradient flow in sparse neural networks and how lottery tickets win.
\newblock \emph{ArXiv}, abs/2010.03533, 2020{\natexlab{b}}.

\bibitem[Evci et~al.(2020{\natexlab{c}})Evci, Ioannou, Keskin, and
  Dauphin]{evci2020gradient}
Evci, U., Ioannou, Y.~A., Keskin, C., and Dauphin, Y.
\newblock Gradient flow in sparse neural networks and how lottery tickets win.
\newblock \emph{arXiv preprint arXiv:2010.03533}, 2020{\natexlab{c}}.

\bibitem[Frankle \& Carbin(2018)Frankle and Carbin]{frankle2018lottery}
Frankle, J. and Carbin, M.
\newblock The lottery ticket hypothesis: Finding sparse, trainable neural
  networks.
\newblock \emph{arXiv preprint arXiv:1803.03635}, 2018.

\bibitem[Frankle et~al.(2019)Frankle, Dziugaite, Roy, and
  Carbin]{frankle2019stabilizing}
Frankle, J., Dziugaite, G.~K., Roy, D.~M., and Carbin, M.
\newblock Stabilizing the lottery ticket hypothesis.
\newblock \emph{arXiv preprint arXiv:1903.01611}, 2019.

\bibitem[Frankle et~al.(2020)Frankle, Dziugaite, Roy, and
  Carbin]{frankle2020pruning}
Frankle, J., Dziugaite, G.~K., Roy, D.~M., and Carbin, M.
\newblock Pruning neural networks at initialization: Why are we missing the
  mark?
\newblock \emph{arXiv preprint arXiv:2009.08576}, 2020.

\bibitem[Gale et~al.(2019)Gale, Elsen, and Hooker]{gale2019state}
Gale, T., Elsen, E., and Hooker, S.
\newblock The state of sparsity in deep neural networks.
\newblock \emph{arXiv preprint arXiv:1902.09574}, 2019.

\bibitem[Glorot \& Bengio(2010)Glorot and Bengio]{Glorot2010UnderstandingTD}
Glorot, X. and Bengio, Y.
\newblock Understanding the difficulty of training deep feedforward neural
  networks.
\newblock In \emph{AISTATS}, 2010.

\bibitem[Han et~al.(2015{\natexlab{a}})Han, Mao, and Dally]{han2015deep}
Han, S., Mao, H., and Dally, W.~J.
\newblock Deep compression: Compressing deep neural networks with pruning,
  trained quantization and huffman coding.
\newblock \emph{arXiv preprint arXiv:1510.00149}, 2015{\natexlab{a}}.

\bibitem[Han et~al.(2015{\natexlab{b}})Han, Pool, Tran, and
  Dally]{han2015learning}
Han, S., Pool, J., Tran, J., and Dally, W.
\newblock Learning both weights and connections for efficient neural network.
\newblock In \emph{Advances in neural information processing systems}, pp.\
  1135--1143, 2015{\natexlab{b}}.

\bibitem[He et~al.(2016)He, Zhang, Ren, and Sun]{He2016DeepRL}
He, K., Zhang, X., Ren, S., and Sun, J.
\newblock Deep residual learning for image recognition.
\newblock \emph{2016 IEEE Conference on Computer Vision and Pattern Recognition
  (CVPR)}, pp.\  770--778, 2016.

\bibitem[He et~al.(2017)He, Zhang, and Sun]{he2017channel}
He, Y., Zhang, X., and Sun, J.
\newblock Channel pruning for accelerating very deep neural networks.
\newblock In \emph{Proceedings of the IEEE International Conference on Computer
  Vision}, pp.\  1389--1397, 2017.

\bibitem[Hinton et~al.(2015{\natexlab{a}})Hinton, Vinyals, and
  Dean]{hinton2015distilling}
Hinton, G., Vinyals, O., and Dean, J.
\newblock Distilling the knowledge in a neural network.
\newblock \emph{arXiv preprint arXiv:1503.02531}, 2015{\natexlab{a}}.

\bibitem[Hinton et~al.(2015{\natexlab{b}})Hinton, Vinyals, and
  Dean]{Hinton2015DistillingTK}
Hinton, G.~E., Vinyals, O., and Dean, J.
\newblock Distilling the knowledge in a neural network.
\newblock \emph{ArXiv}, abs/1503.02531, 2015{\natexlab{b}}.

\bibitem[Howard et~al.(2017)Howard, Zhu, Chen, Kalenichenko, Wang, Weyand,
  Andreetto, and Adam]{howard2017mobilenets}
Howard, A.~G., Zhu, M., Chen, B., Kalenichenko, D., Wang, W., Weyand, T.,
  Andreetto, M., and Adam, H.
\newblock Mobilenets: Efficient convolutional neural networks for mobile vision
  applications.
\newblock \emph{arXiv preprint arXiv:1704.04861}, 2017.

\bibitem[Hubara et~al.(2018)Hubara, Courbariaux, Soudry, El-Yaniv, and
  Bengio]{quantHilton2018}
Hubara, I., Courbariaux, M., Soudry, D., El-Yaniv, R., and Bengio, Y.
\newblock Quantized neural networks: Training neural networks with low
  precision weights and activations.
\newblock \emph{Journal of Machine Learning Research}, 18\penalty0
  (187):\penalty0 1--30, 2018.
\newblock URL \url{http://jmlr.org/papers/v18/16-456.html}.

\bibitem[Izmailov et~al.(2018)Izmailov, Podoprikhin, Garipov, Vetrov, and
  Wilson]{izmailov2018averaging}
Izmailov, P., Podoprikhin, D., Garipov, T., Vetrov, D., and Wilson, A.~G.
\newblock Averaging weights leads to wider optima and better generalization.
\newblock \emph{arXiv preprint arXiv:1803.05407}, 2018.

\bibitem[Jean \& Wang(1994)Jean and Wang]{jean1994weight}
Jean, J.~S. and Wang, J.
\newblock Weight smoothing to improve network generalization.
\newblock \emph{IEEE Transactions on neural networks}, 5\penalty0 (5):\penalty0
  752--763, 1994.

\bibitem[Leclerc \& Madry(2020)Leclerc and Madry]{leclerc2020two}
Leclerc, G. and Madry, A.
\newblock The two regimes of deep network training.
\newblock \emph{arXiv preprint arXiv:2002.10376}, 2020.

\bibitem[LeCun et~al.(1989)LeCun, Denker, and Solla]{LeCun1989OptimalBD}
LeCun, Y., Denker, J.~S., and Solla, S.~A.
\newblock Optimal brain damage.
\newblock In \emph{NIPS}, 1989.

\bibitem[LeCun et~al.(1990)LeCun, Denker, and Solla]{lecun1990optimal}
LeCun, Y., Denker, J.~S., and Solla, S.~A.
\newblock Optimal brain damage.
\newblock In \emph{Advances in neural information processing systems}, pp.\
  598--605, 1990.

\bibitem[Lee et~al.(2018)Lee, Ajanthan, and Torr]{lee2018snip}
Lee, N., Ajanthan, T., and Torr, P.~H.
\newblock Snip: Single-shot network pruning based on connection sensitivity.
\newblock \emph{arXiv preprint arXiv:1810.02340}, 2018.

\bibitem[Lee et~al.(2019)Lee, Ajanthan, Gould, and Torr]{lee2019signal}
Lee, N., Ajanthan, T., Gould, S., and Torr, P.~H.
\newblock A signal propagation perspective for pruning neural networks at
  initialization.
\newblock \emph{arXiv preprint arXiv:1906.06307}, 2019.

\bibitem[Li et~al.(2016)Li, Kadav, Durdanovic, Samet, and Graf]{li2016pruning}
Li, H., Kadav, A., Durdanovic, I., Samet, H., and Graf, H.~P.
\newblock Pruning filters for efficient convnets.
\newblock \emph{arXiv preprint arXiv:1608.08710}, 2016.

\bibitem[Li et~al.(2017{\natexlab{a}})Li, Xu, Taylor, and
  Goldstein]{loss-landscape}
Li, H., Xu, Z., Taylor, G., and Goldstein, T.
\newblock Visualizing the loss landscape of neural nets.
\newblock \emph{CoRR}, abs/1712.09913, 2017{\natexlab{a}}.
\newblock URL \url{http://arxiv.org/abs/1712.09913}.

\bibitem[Li et~al.(2017{\natexlab{b}})Li, Xu, Taylor, Studer, and
  Goldstein]{li2017visualizing}
Li, H., Xu, Z., Taylor, G., Studer, C., and Goldstein, T.
\newblock Visualizing the loss landscape of neural nets.
\newblock \emph{arXiv preprint arXiv:1712.09913}, 2017{\natexlab{b}}.

\bibitem[Liu et~al.(2021)Liu, Yin, Mocanu, and Pechenizkiy]{liu2021we}
Liu, S., Yin, L., Mocanu, D.~C., and Pechenizkiy, M.
\newblock Do we actually need dense over-parameterization? in-time
  over-parameterization in sparse training.
\newblock \emph{arXiv preprint arXiv:2102.02887}, 2021.

\bibitem[Liu et~al.(2017)Liu, Li, Shen, Huang, Yan, and Zhang]{liu2017learning}
Liu, Z., Li, J., Shen, Z., Huang, G., Yan, S., and Zhang, C.
\newblock Learning efficient convolutional networks through network slimming.
\newblock In \emph{Proceedings of the IEEE international conference on computer
  vision}, pp.\  2736--2744, 2017.

\bibitem[Liu et~al.(2019)Liu, Sun, Zhou, Huang, and Darrell]{liu2018rethinking}
Liu, Z., Sun, M., Zhou, T., Huang, G., and Darrell, T.
\newblock Rethinking the value of network pruning.
\newblock In \emph{International Conference on Learning Representations}, 2019.

\bibitem[Misra(2019)]{misra2019mish}
Misra, D.
\newblock Mish: A self regularized non-monotonic neural activation function.
\newblock \emph{arXiv preprint arXiv:1908.08681}, 4:\penalty0 2, 2019.

\bibitem[Mocanu et~al.(2018)Mocanu, Mocanu, Stone, Nguyen, Gibescu, and
  Liotta]{mocanu2018scalable}
Mocanu, D.~C., Mocanu, E., Stone, P., Nguyen, P.~H., Gibescu, M., and Liotta,
  A.
\newblock Scalable training of artificial neural networks with adaptive sparse
  connectivity inspired by network science.
\newblock \emph{Nature communications}, 9\penalty0 (1):\penalty0 1--12, 2018.

\bibitem[Molchanov et~al.(2017)Molchanov, Ashukha, and
  Vetrov]{molchanov2017variational}
Molchanov, D., Ashukha, A., and Vetrov, D.
\newblock Variational dropout sparsifies deep neural networks.
\newblock In \emph{International Conference on Machine Learning}, pp.\
  2498--2507. PMLR, 2017.

\bibitem[Molchanov et~al.(2019)Molchanov, Mallya, Tyree, Frosio, and
  Kautz]{molchanov2019importance}
Molchanov, P., Mallya, A., Tyree, S., Frosio, I., and Kautz, J.
\newblock Importance estimation for neural network pruning.
\newblock In \emph{Proceedings of the IEEE Conference on Computer Vision and
  Pattern Recognition}, pp.\  11264--11272, 2019.

\bibitem[M{\"u}ller et~al.(2019)M{\"u}ller, Kornblith, and
  Hinton]{muller2019does}
M{\"u}ller, R., Kornblith, S., and Hinton, G.
\newblock When does label smoothing help?
\newblock \emph{arXiv preprint arXiv:1906.02629}, 2019.

\bibitem[M{\"{u}}ller et~al.(2019)M{\"{u}}ller, Kornblith, and
  Hinton]{label-smoothening}
M{\"{u}}ller, R., Kornblith, S., and Hinton, G.~E.
\newblock When does label smoothing help?
\newblock \emph{CoRR}, abs/1906.02629, 2019.
\newblock URL \url{http://arxiv.org/abs/1906.02629}.

\bibitem[Nair \& Hinton(2010)Nair and Hinton]{nair2010rectified}
Nair, V. and Hinton, G.~E.
\newblock Rectified linear units improve restricted boltzmann machines.
\newblock In \emph{Icml}, 2010.

\bibitem[Petzka et~al.(2019)Petzka, Adilova, Kamp, and
  Sminchisescu]{Petzka2019ARF}
Petzka, H., Adilova, L., Kamp, M., and Sminchisescu, C.
\newblock A reparameterization-invariant flatness measure for deep neural
  networks.
\newblock \emph{ArXiv}, abs/1912.00058, 2019.

\bibitem[Rahaman et~al.(2019)Rahaman, Baratin, Arpit, Draxler, Lin, Hamprecht,
  Bengio, and Courville]{rahaman2019spectral}
Rahaman, N., Baratin, A., Arpit, D., Draxler, F., Lin, M., Hamprecht, F.,
  Bengio, Y., and Courville, A.
\newblock On the spectral bias of neural networks.
\newblock In \emph{International Conference on Machine Learning}, pp.\
  5301--5310. PMLR, 2019.

\bibitem[Ramachandran et~al.(2017)Ramachandran, Zoph, and
  Le]{ramachandran2017searching}
Ramachandran, P., Zoph, B., and Le, Q.~V.
\newblock Searching for activation functions.
\newblock \emph{arXiv preprint arXiv:1710.05941}, 2017.

\bibitem[Savarese et~al.(2020)Savarese, Silva, and Maire]{savarese2020winning}
Savarese, P., Silva, H., and Maire, M.
\newblock Winning the lottery with continuous sparsification.
\newblock In \emph{NeurIPS}, 2020.

\bibitem[Simonyan \& Zisserman(2014)Simonyan and Zisserman]{simonyan2014very}
Simonyan, K. and Zisserman, A.
\newblock Very deep convolutional networks for large-scale image recognition.
\newblock \emph{arXiv preprint arXiv:1409.1556}, 2014.

\bibitem[Su et~al.(2020)Su, Chen, Cai, Wu, Gao, Wang, and Lee]{su2020sanity}
Su, J., Chen, Y., Cai, T., Wu, T., Gao, R., Wang, L., and Lee, J.~D.
\newblock Sanity-checking pruning methods: Random tickets can win the jackpot.
\newblock \emph{arXiv preprint arXiv:2009.11094}, 2020.

\bibitem[Sung et~al.(2021)Sung, Nair, and Raffel]{sung2021training}
Sung, Y.-L., Nair, V., and Raffel, C.
\newblock Training neural networks with fixed sparse masks.
\newblock In Beygelzimer, A., Dauphin, Y., Liang, P., and Vaughan, J.~W.
  (eds.), \emph{Advances in Neural Information Processing Systems}, 2021.
\newblock URL \url{https://openreview.net/forum?id=Uwh-v1HSw-x}.

\bibitem[Szegedy et~al.(2016)Szegedy, Vanhoucke, Ioffe, Shlens, and
  Wojna]{szegedy2016rethinking}
Szegedy, C., Vanhoucke, V., Ioffe, S., Shlens, J., and Wojna, Z.
\newblock Rethinking the inception architecture for computer vision.
\newblock In \emph{Proceedings of the IEEE conference on computer vision and
  pattern recognition}, pp.\  2818--2826, 2016.

\bibitem[Tanaka et~al.(2020)Tanaka, Kunin, Yamins, and
  Ganguli]{tanaka2020pruning}
Tanaka, H., Kunin, D., Yamins, D.~L., and Ganguli, S.
\newblock Pruning neural networks without any data by iteratively conserving
  synaptic flow.
\newblock \emph{arXiv preprint arXiv:2006.05467}, 2020.

\bibitem[Tessera et~al.(2021)Tessera, Hooker, and Rosman]{tessera2021keep}
Tessera, K.-a., Hooker, S., and Rosman, B.
\newblock Keep the gradients flowing: Using gradient flow to study sparse
  network optimization.
\newblock \emph{arXiv preprint arXiv:2102.01670}, 2021.

\bibitem[Vaswani et~al.(2017)Vaswani, Shazeer, Parmar, Uszkoreit, Jones, Gomez,
  Kaiser, and Polosukhin]{Vaswani2017AttentionIA}
Vaswani, A., Shazeer, N.~M., Parmar, N., Uszkoreit, J., Jones, L., Gomez,
  A.~N., Kaiser, L., and Polosukhin, I.
\newblock Attention is all you need.
\newblock \emph{ArXiv}, abs/1706.03762, 2017.

\bibitem[Wang et~al.(2020)Wang, Zhang, and Grosse]{wang2020picking}
Wang, C., Zhang, G., and Grosse, R.
\newblock Picking winning tickets before training by preserving gradient flow.
\newblock \emph{arXiv preprint arXiv:2002.07376}, 2020.

\bibitem[Wen et~al.(2016)Wen, Wu, Wang, Chen, and Li]{wen2016learning}
Wen, W., Wu, C., Wang, Y., Chen, Y., and Li, H.
\newblock Learning structured sparsity in deep neural networks.
\newblock In \emph{Advances in neural information processing systems}, pp.\
  2074--2082, 2016.

\bibitem[Yao et~al.(2020)Yao, Gholami, Keutzer, and
  Mahoney]{Yao2020PyHessianNN}
Yao, Z., Gholami, A., Keutzer, K., and Mahoney, M.~W.
\newblock Pyhessian: Neural networks through the lens of the hessian.
\newblock \emph{2020 IEEE International Conference on Big Data (Big Data)},
  pp.\  581--590, 2020.

\bibitem[Zhu et~al.(2021)Zhu, Ni, Xu, Kong, Huang, and
  Goldstein]{Zhu2021GradInitLT}
Zhu, C., Ni, R., Xu, Z., Kong, K., Huang, W.~R., and Goldstein, T.
\newblock Gradinit: Learning to initialize neural networks for stable and
  efficient training.
\newblock \emph{ArXiv}, abs/2102.08098, 2021.

\end{thebibliography}
\bibliographystyle{icml2022}


\end{document}